\documentclass[10pt,twocolumn,letterpaper]{article}

\usepackage{cvpr}              
\usepackage[accsupp]{axessibility}

\usepackage{newfloat}
\usepackage{flushend}
\usepackage{listings}
\usepackage{makecell}
\usepackage{amsmath}
\DeclareCaptionStyle{ruled}{labelfont=normalfont,labelsep=colon,strut=off} 
\usepackage{graphicx}
\usepackage{subcaption}

\definecolor{cvprblue}{rgb}{0.21,0.49,0.74}
\usepackage[pagebackref,breaklinks,colorlinks,allcolors=cvprblue]{hyperref}
\usepackage{orcidlink}

\def\paperID{9673} 
\def\confName{CVPR}
\def\confYear{2025}

\title{PhD: A ChatGPT-\underline{P}rompted Visual \underline{h}allucination Evaluation \underline{D}ataset}

\author{Jiazhen Liu$^{1}$\thanks{J. Liu is currently pursuing his PhD at HKUST.}  \orcidlink{0000-0003-0584-4571}, Yuhan Fu$^{1,2}$, Ruobing Xie$^{2}$, Runquan Xie$^{2}$, Xingwu Sun$^{2}$, Fengzong Lian$^{2}$, \\ Zhanhui Kang$^{2}$, and Xirong Li$^{1}$\thanks{Corresponding author (xirong@ruc.edu.cn)} \orcidlink{0000-0002-0220-8310}\\
\\
$^{1}$Renmin University of China \\ 
$^{2}$Machine Learning Platform Department, Tencent \\
\tt\small\href{https://github.com/jiazhen-code/PhD}{https://github.com/jiazhen-code/PhD}
}

\newcommand{\ra}[1]{\renewcommand{\arraystretch}{#1}}
\usepackage{adjustbox}
\usepackage{bibentry}

\usepackage{pifont}
\newcommand{\xmark}{\ding{55}}%

\newcommand{\specialcell}[2][c]{\begin{tabular}[#1]{@{}l@{}}#2\end{tabular}}

\usepackage{multirow} 

\usepackage{colortbl}
\usepackage{pgfplotstable}
\usepackage{siunitx}  

\definecolor{blue}{RGB}{230, 240, 255} 
\definecolor{red}{RGB}{255, 150, 150} 

\definecolor{highcolor}{RGB}{230, 240, 255} 
\definecolor{lowcolor}{RGB}{255, 150, 150} 

\newcommand{\applyGradient}[1]{%
    \begingroup
    \pgfmathsetmacro{\value}{((#1 * 100 - 35) / (90-35)) * 100} 
    \edef\tempcolor{\noexpand\cellcolor{lowcolor!\value!highcolor}}%
    \tempcolor #1
    \endgroup
}

\newcommand{\phd}{\texttt{PhD}}

\begin{document}
\maketitle
\begin{abstract}
Multimodal Large Language Models (MLLMs) hallucinate, resulting in an emerging topic of visual hallucination evaluation (VHE). This paper contributes a ChatGPT-\textbf{P}rompted visual \textbf{h}allucination  evaluation \textbf{D}ataset (\emph{PhD}) for objective VHE at a large scale.  The essence of VHE is to ask an MLLM questions about specific images to assess its susceptibility to hallucination. Depending on what to ask (objects, attributes, sentiment, etc.) and how the questions are asked, we structure PhD along two dimensions, \ie task and mode. Five visual recognition tasks, ranging from low-level (object  / attribute recognition) to middle-level (sentiment / position recognition and counting), are considered. Besides a normal visual QA mode, which we term \phd-base, PhD also asks questions with specious context (\phd-sec) or with incorrect context (\phd-icc), or with AI-generated counter common sense images (\phd-ccs). We construct PhD by a ChatGPT-assisted semi-automated pipeline, encompassing four pivotal modules: task-specific hallucinatory item (hitem) selection, hitem-embedded question generation, specious / incorrect context generation, and counter-common-sense (CCS) image generation. With over 14k daily images, 750 CCS images and 102k VQA triplets in total, PhD reveals considerable variability in MLLMs' performance across various modes and tasks, offering valuable insights into the nature of hallucination. As such, PhD stands as a potent tool not only for VHE but may also play a significant role in the refinement of MLLMs.
\end{abstract}

\section{Introduction}



\begin{figure}[!t]
    \centering
    
    \begin{subfigure}[t]{0.47\textwidth}
        \centering
        \includegraphics[width=\textwidth,height=3cm,keepaspectratio]{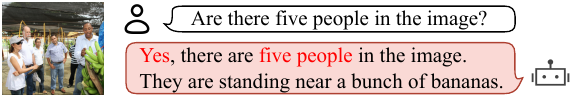}
        \caption{Hallucination cause I: Visual ambiguity (MLLM: LLaVA-1.5) \cite{liu2023improvedllava}}
        \label{fig:subfigB}
    \end{subfigure}
    \par\medskip  
    
    \begin{subfigure}[t]{0.47\textwidth}
        \centering
        \includegraphics[width=\textwidth,height=3cm,keepaspectratio]{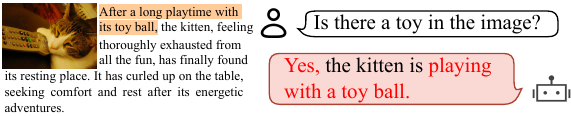}
        \caption{Hallucination cause II: Inconsistency in multi-modal input}
        \label{fig:subfigC}
    \end{subfigure}
    \par\medskip  
    
    \begin{subfigure}[t]{0.47\textwidth}
        \centering
        \includegraphics[width=\textwidth,height=3cm,keepaspectratio]{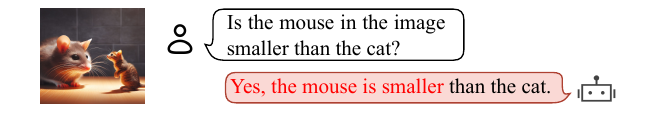}
        \caption{Hallucination cause III: Counter-common-sense content}
        \label{fig:subfigD}
    \end{subfigure}
    \par\medskip  

    \begin{subfigure}[t]{0.47\textwidth}
        \centering
        \includegraphics[width=\textwidth]{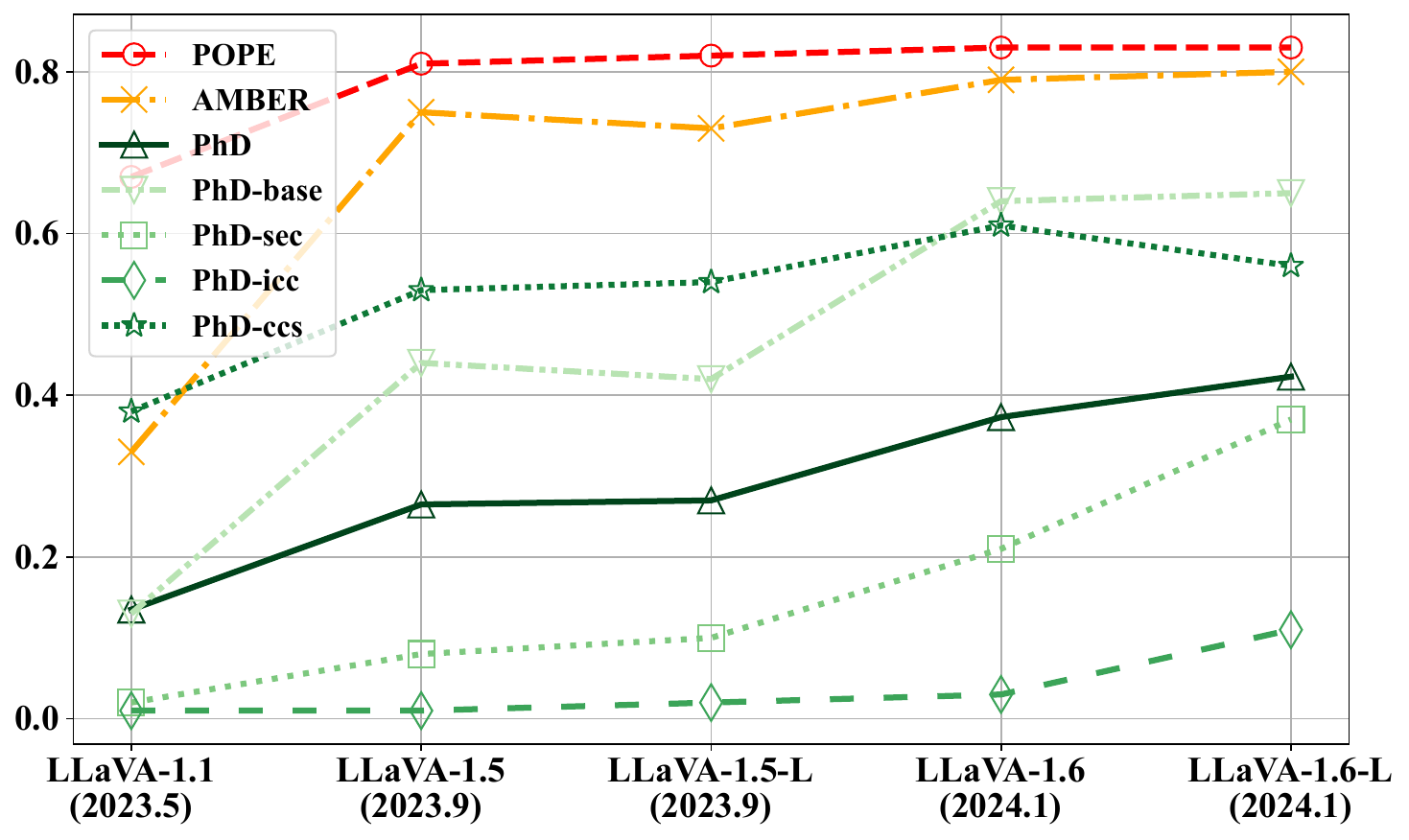}
        \caption{Performance curves of the LLaVA series on two public VHE datasets (POPE \cite{pope}) and AMBER \cite{amber}) and the proposed \phd~ dataset. 
        }
        \label{fig:develop}
    \end{subfigure}

    \caption{
    \textbf{Illustrations of three major causes of an MLLM's visual hallucination and its evaluation}.
    This paper contributes \phd, a  binary VQA-based VHE benchmark, much larger and more challenging than its predecessors. 
    In particular, it has four evaluation modes that \emph{explicitly} measure an MLLM's performance \wrt the three causes, \ie \phd-base for cause I, \phd-sec and \phd-icc for cause II and \phd-ccs for cause III.  
    }
    \label{fig:intro}
\end{figure}

Using a specific large language model (LLM) as its kernel, a multi-modal LLM (MLLM), exemplified by LLaVA \cite{liu2023improvedllava}, Qwen-VL \cite{bai2023qwenvl} and MiniGPT-v2 \cite{chen2023minigptv2} can now tackle a wide range of computer vision tasks in a unified visual-question-answering (VQA) manner. As LLMs are known to hallucinate \cite{zhang2023siren, xu2024hallucination, li-etal-2024-dawn}, it is not surprising that MLLMs have visual hallucination, generating fabricated interpretation of the given visual content, see \cref{fig:intro}. Considering the rapidly growing use of MLLMs in varied scenarios, a comprehensive visual hallucination evaluation (VHE) is crucial. This paper develops a new dataset for VHE.

VHE essentially involves posing a number of visual questions to an MLLM \cite{pope,liu2023hallusionbench}.  
A question of this kind shall include a \underline{h}allucinatory \underline{item} (\textbf{hitem}), in the form of  a specific word or phrase, that induces the MLLM to generate a response discordant with the provided visual content. 
As the model typically has a strong visual recognition ability, \emph{how to identify an appropriate hitem} and accordingly generate a proper question is nontrivial. Both the hitem and the question depend on the visual recognition task being considered. As shown in \cref{tab:taxonomy}, \textbf{we target at objective VHE in} the context of \textbf{low-level}   (\emph{object} / \emph{attribute}) to \textbf{middle-level} (\emph{sentiment} / \emph{position} / \emph{counting}) visual recognition \textbf{tasks}. 
Such a target is chosen due to the following considerations. MLLMs generally work well for these tasks, so their erroneous responses can be largely attributed to their hallucinations instead of their incapability, \eg asking a generic MLLM to read pathology images \cite{OmniMedVQA}. Meanwhile, a binary VQA based objective evaluation is more budget friendly and thus more suited for VHE at a large scale.

\definecolor{lightyellow}{RGB}{255, 255, 204}
\begin{table}[!htb]
    \centering
    \begin{adjustbox}{max width=\linewidth}
    \begin{tabular}{@{}ll|l@{}}
    \toprule
\textbf{Vision tasks}  & \textbf{Objective evaluation} &	\textbf{Subjective evaluation} \\
     \midrule
\specialcell{\textit{Low-/middle-level}\\ \textit{visual recognition}} & \cellcolor{lightyellow}\specialcell{+ POPE, EMNLP'23 \cite{pope}\\ + AMBER, arXiv'23 \cite{amber} \\+ CIEM, ITIF'23 \cite{hu2023ciem} \\ + NOPE, ALVR'24 \cite{nope}\\ + ROME, EMNLP'23 \cite{rome} \\ + \texttt{PhD} (\emph{this paper})} &	\specialcell{+ FAITHSCORE, EMNLP'24 \cite{faithscore}\\ + HaELM, arXiv'23 \cite{haelm} \\ + M-HalDetect, AAAI'24 \cite{mhaldetect} \\ + GAVIE, ICLR'24 \cite{liu2024mitigating} }	\\
      \cmidrule{2-3}
\specialcell{\textit{High-level}\\ \textit{visual reasoning}}	& \specialcell{+ MMMU, CVPR'24 \cite{yue2023mmmu} \\ + VLind-Bench, NAACL'25 \cite{lee2024vlind}} & \specialcell{+ HallusionBench, CVPR'24 \cite{liu2023hallusionbench} \\ + Bingo, arXiv'23 \cite{bingo} \\ + IllusionVQA, COLM'24 \cite{shahgir2024illusionvqa} \\ + WHOOPS!, ICCV'23\cite{bitton2023breaking}}\\
 \bottomrule
    \end{tabular}
    \end{adjustbox}
    \caption{\textbf{Taxonomy of VHE benchmarks}. Our \texttt{PhD} benchmark performs an objective evaluation of MLLMs' hallucinations when they address  visual recognition tasks ranging from low-level, \ie \emph{object / attribute recognition} to middle-level, \ie \emph{sentiment / positional recognition} and \emph{counting}.}
    \label{tab:taxonomy}
\end{table}

Pioneered by POPE \cite{pope}, good attempts exist in objective VHE 
\cite{nope,hu2023ciem,amber,rome}. 
In these valuable datasets, 
hitem selection is largely untouched. POPE and ROME \cite{rome} select their hitems fully based on label co-occurrence in training data, AMBER \cite{amber} relies on manual annotation, whilst  hitem selection is not considered in NOPE \cite{nope} and CIEM \cite{hu2023ciem}, see \cref{tab:comp-datasets}. Hence, \textbf{there lacks an explicit link between hitem selection} (and subsequent VQA triplets construction) and \textbf{major causes of an MLLM's visual hallucination}. As models rapidly evolve, the performance on these datasets quickly reaches saturation, see \cref{fig:develop}.

Analyzing an MLLM's typical dataflow of answering a visual question, we see three major causes of visual hallucination: I) visual ambiguity, II) inconsistency in multi-modal input and III) counter-common-sense (CCS) content, 
see \cref{fig:intro}.
Firstly, the MLLM extracts tokenized visual features from a given image using a ViT based encoder. 
Recent studies \cite{tong2024eyes, zhang2024contrasting} show that the features tend to be high level, lacking sufficient details for fine-grained tasks such as \emph{counting}. Secondly, the visual features, after vision-to-language adaptation, are  mixed with the features of the associated textual prompt to form a multi-modal input to the LLM kernel. The LLM, pre-trained extensively on textual data, inevitably favors the textual part of the multi-modal input. Hence, when there is inconsistency between the visual and textual information, the former is more likely to be overruled. Lastly, at the decoding stage, the LLM might heavily rely on its internal (world) knowledge, ignoring the visual content especially when the content (showing \emph{a mouse much larger than a cat}) contradicts the common sense. \textbf{Our new dataset is developed with a close link to the three causes}.

\begin{table}[!tbhp]
\fontsize{14pt}{15pt}\selectfont
    \ra{1.35}
        \centering
        \begin{adjustbox}{max width=\linewidth}
        \begin{tabular}{@{}lrrrrrr@{}}
            \toprule
            \textbf{Dataset} & \textbf{Daily images} & \textbf{CCS images} & \textbf{Hitems} & \textbf{Contexts} & \textbf{VQA triplets} & \textbf{Tasks} \\
            \hline
            POPE & 500 & \xmark & 80 & \xmark & 3,000 & Obj. \\
            \hline
            NOPE* &	unknown &	\xmark &	unknown &	\xmark	&  32,701 &	Obj. \\
            \hline
            CIEM* &	4,929 &	\xmark &	unknown &	\xmark & 72,941 &	 Obj. / Attr. / Pos. \\
            \hline
            AMBER & 1,004 & \xmark & 687 & \xmark & 14,216 & \makecell[r]{Obj. / Attr./ Pos.\\ / Count.} \\
            \hline
            ROME & \xmark & 1,563 & 118 & \xmark & 1,563 & \makecell[r]{Attr./ Pos.} \\
            \hline
            PhD & 14,648 & 750 & 1,452 & 33,688 & 102,564 & \makecell[r]{Obj / Attr. / Pos.\\ / Sent. / Count.} \\
            \bottomrule
        \end{tabular}
        \end{adjustbox}
        \caption{\textbf{PhD versus its predecessors}. * indicates private dataset. 
        }
        \label{tab:comp-datasets}

\end{table}

The new dataset is constructed by adapting TDIUC, a popular multi-task VQA dataset \cite{tdiuc}, with a ChatGPT-assisted semi-automated pipeline, see \cref{fig:whole}. In particular, by prompting ChatGPT, we select diverse and visually challenging hitems in an \emph{image-specific} and \emph{task-specific} manner, with minimal human involvement primarily spent on verifying ChatGPT-generated results. The selected hitems are then automatically embedded into visual questions, specious context, and incorrect context, all generated by instructing ChatGPT. Moreover, we expand the daily image set with counter common sense (CCS) images, obtained by prompting AIGC tools with ChatGPT-generated CCS descriptions, \eg ``\emph{trees growing
underwater}'' and ``\emph{a car with square-shaped wheels}''. The dataset is dubbed as \phd~ (ChatGPT \underline{P}rompted visual  \underline{h}allucination evaluation  \underline{D}ataset).
Depending on what image (daily or CCS) is used and whether a specific context precedes a question, \phd~supports four evaluation modes: \phd-base (questions about daily images w/o context), \phd-sec (\phd-base plus specious context), \phd-icc ( \phd-base plus incorrect context), and \phd-ccs (questions about CCS images). 

To sum up, our major contributions are as follows: \\
%
    $\bullet$ We introduce \texttt{PhD}, a dataset with four evaluation modes across five visual recognition tasks, developed with a close link to the three major causes of MLLM visual hallucination. 
    Information on hallucinatory items (hitems) is provided per sample, enabling in-depth analytics to uncover the  causes in more detail.   \\ 
    $\bullet$  We offer a ChatGPT-assisted semi-automatic pipeline for dataset construction, 
    with minimal human involvement, primarily focused on verifying the generated results. With 14,648 daily images, 750 CCS images and 102,564 VQA triplets in total, \phd~ is the largest of its kind. \\ 
    $\bullet$  We conduct an extensive evaluation with 15 open-source MLLMs, 3 proprietary MLLMs, and 2 hallucination mitigation methods, showing the viability of \phd~for VHE in varied manners including overall, mode-oriented, task-oriented, and model-wise zoom-in. The evaluation not only reveals inter-model performance divergence, but may also help developers of a specific MLLM to prioritize their efforts in refining the model.   

\section{Related Work}

Due to the increasing importance of VHE, new  benchmarks are being actively developed. Depending on what vision tasks they focus on and how their evaluation is executed, we categorize existing achievements along two dimensions, \ie \emph{task} and \emph{evaluation}, see  \cref{tab:taxonomy}. Concerning the \emph{task} dimension, low- and middle-level visual recognition tasks, ranging from object / attribute recognition to sentiment / positional recognition and counting, assess an MLLM's basic visual skills. High-level visual reasoning is more domain-knowledge intensive, typically covering image-based math question solving, geography information understanding, meme interpretation, historical or folkloric contexts, \etc \cite{liu2023hallusionbench, lee2024vlind, bitton2023breaking, bingo}. As for the \emph{evaluation} dimension, objective evaluation refers to objectively comparing the model's output with ground truth,  mostly in the form of Yes/No answers \cite{pope, hu2023ciem, amber, rome}. By contrast, subjective evaluation requires humans or LLMs to assess the model's output \cite{liu2023hallusionbench, bingo, bitton2023breaking}. The proposed \texttt{PhD}, focusing on low-/middle-level visual recognition \emph{and} objective evaluation, belongs to the second quadrant of the taxonomy. In what follows, we briefly review peer benchmarks, \ie POPE \cite{pope}, ROME \cite{rome}, NOPE \cite{nope}, CIEM \cite{hu2023ciem}, and AMBER \cite{amber}, in this quadrant and clarify our novelty accordingly.

POPE is probably the first dataset to evaluate object hallucination \cite{pope}. Given a specific MS-COCO image with object labels, POPE selects an adversarial object frequently co-occurring with the current labels. A binary question is then formed by filling out a predefined template with the selected object.  
Such co-occurrence based hitem selection is not \emph{image-specific} by definition. 
Trivial hitems might thus be picked up, \eg ``car'' selected for an indoor image labeled with ``person'', as the two objects often co-occur. 
For CCS image generation, ROME forms CCS descriptions by choosing attribute values having the lowest co-occurrence with a given object according to the Visual Genome dataset \cite{rome}. As low occurrence is not necessarily CCS, some of ROME images are indeed normal.  
NOPE \cite{nope} and CIEM \cite{hu2023ciem} simply bypass hitem selection by asking an LLM to generate questions conditioned on the image captions (also from MS-COCO) and pre-specified answers. 

To select hitems in an image-specific manner,  AMBER  resorts to fully manual annotation \cite{amber}. Manual labeling is costly, while an annotator's personal vocabulary is relatively limited. All this makes it difficult to scale up \wrt the amount of test images and the number of distinct hitems.

In comparison, \texttt{PhD}, constructed by a ChatGPT-assisted semi-automatic pipeline (\cref{sec:roadmap}), is much larger (\cref{tab:comp-datasets}) and more challenging (\cref{fig:develop}). With its unique \emph{mode-task} structure, the new dataset enables a novel, structured and zoom-in understanding of inter-model difference.

\section{Our Roadmap to PhD} \label{sec:roadmap}

\begin{figure*}[!th]
    \centering
        \subfloat[Hitem-embedded question / context generation for MS-COCO daily images]{
            \includegraphics[width=0.69\textwidth]{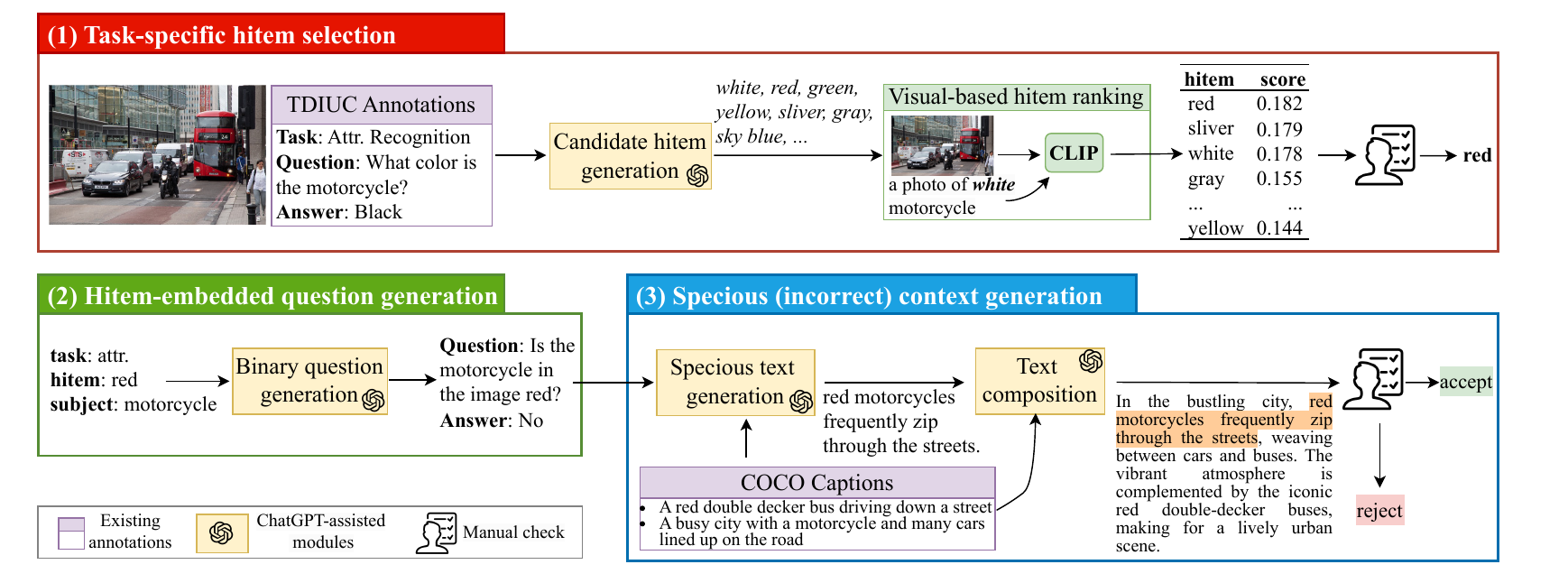}
            \label{fig:hitem}
        } 
        \subfloat[QAs for AI-generated CCS images]{
            \includegraphics[width=0.28\textwidth]{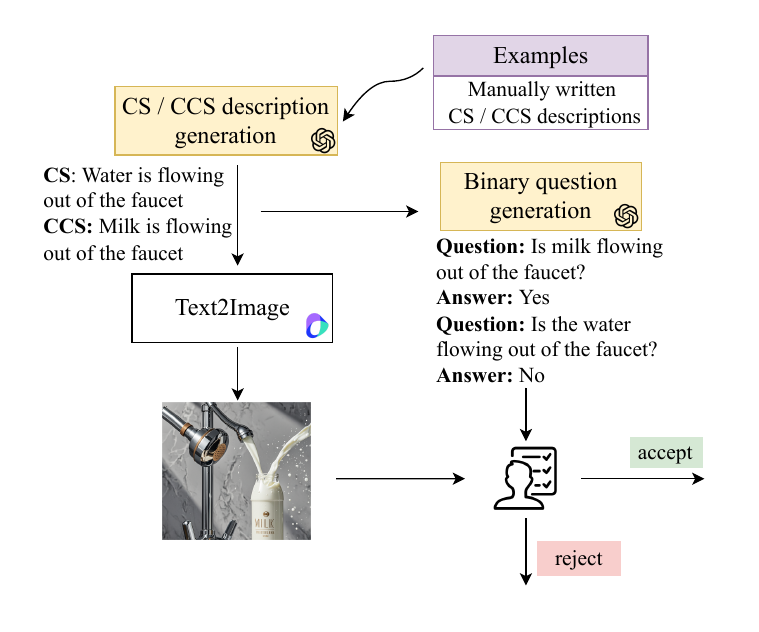}
            \label{fig:aigc}
        } \vskip\baselineskip
        \subfloat[Showcase: Daily images, questions and optional contexts]{
            \includegraphics[width=0.69\textwidth]{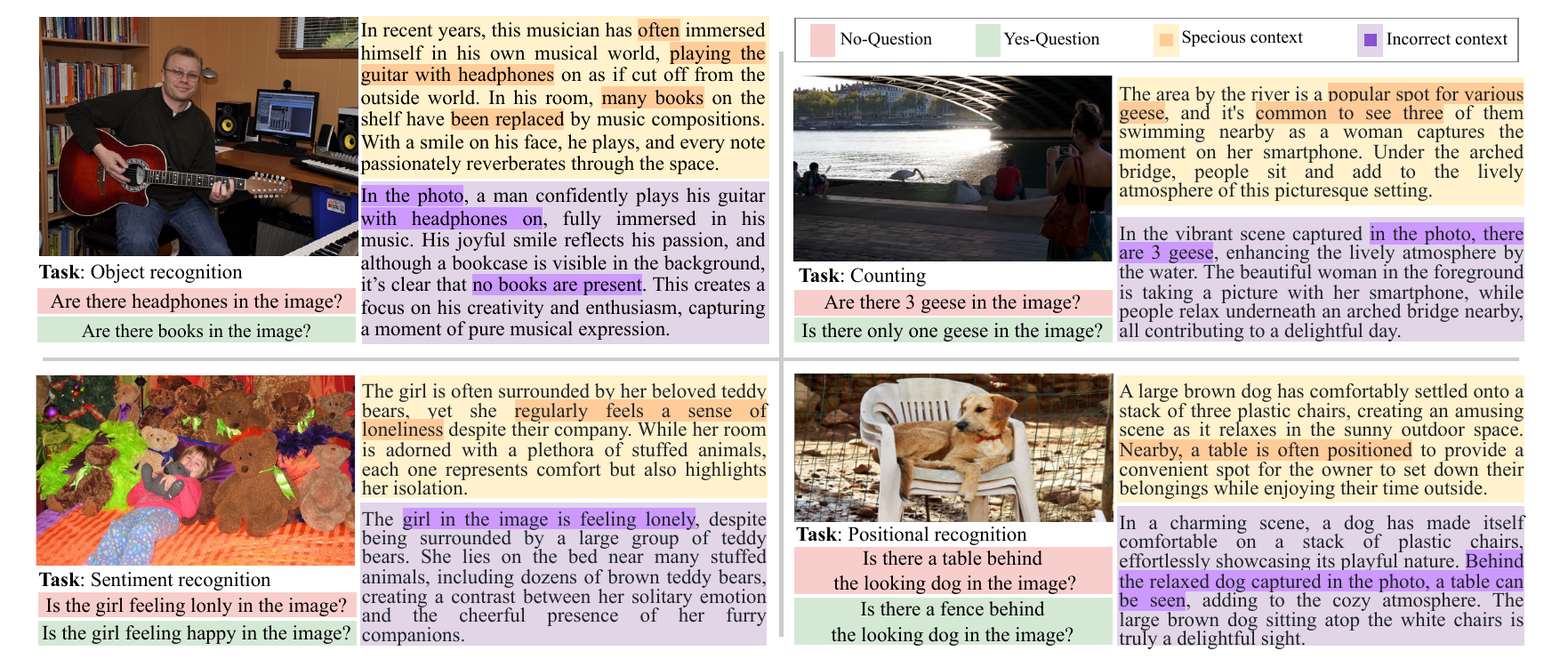}
            \label{fig:context_show}
        } 
        \subfloat[Showcase: CCS images and questions]{
            \includegraphics[width=0.28\textwidth]{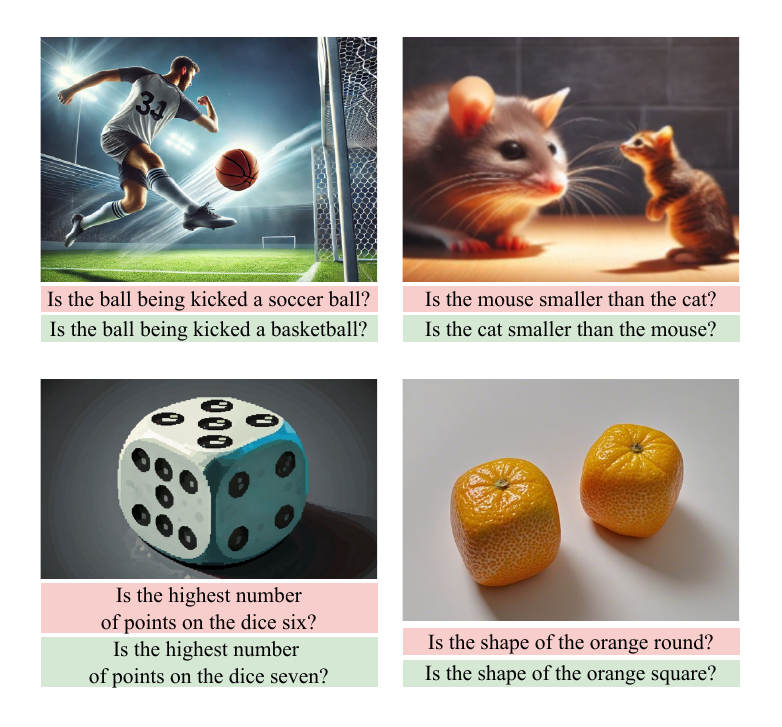}
            \label{fig:aigc_show}
        }
    \caption{\textbf{Proposed semi-automatic pipeline for \phd~ construction}. We use \texttt{ChaptGPT} (GPT-4o mini) to generate hitem-embedded questions / contexts for daily images, and \texttt{Doubao} and \texttt{DALL-E3} for generating CCS images. Depending on what image (daily or CCS) is used and whether a specific context precedes a question, PhD supports four evaluation modes: \textbf{\phd-base}, \ie questions about daily images w/o context, \textbf{\phd-sec}, \ie \phd-base plus specious context, \textbf{\phd-icc}, \ie \phd-base plus incorrect context, and \textbf{\phd-ccs}, \ie questions about CCS images. 
    By adapting TDIUC annotations, \phd~ supports binary VQA \wrt five visual recognition tasks  including \emph{object / attribute / sentiment / positional} recognition and \emph{counting}.
    With 20 mode-task combinations in total, PhD enables a comprehensive VHE. 
    }
    \label{fig:whole}
\end{figure*}

As MLLMs perform visual recognition in a VQA manner, a VQA sample for VHE naturally depends on the visual recognition task in consideration. We depart from TDIUC \cite{tdiuc}, a large-scale VQA dataset \wrt five tasks including \emph{object} / \emph{attribute} / \emph{sentiment} / \emph{positional} recognition and \emph{counting}. Note that the images in TDIUC are sourced from MS-COCO \cite{lin2014microsoft}, which plausibly have been seen by MLLMs in their development stage. As such, our adoption of TDIUC makes \emph{on-purpose} data leakage: an MLLM's erroneous response \wrt a seen image can now be more safely attributed to its hallucination other than its incapability in visual recognition, say asking LLaVA to recognize glaucoma from color fundus photographs \cite{wei2025funbench}. We construct PhD by adapting the TDIUC annotations with a ChatGPT\footnote{We use GPT-4o mini, released on 2024-05-13.}-assisted semi-automated pipeline, see \cref{fig:whole}.


In order to compose a proper question that effectively makes an MLLM hallucinate about a given image, a hitem has to be first identified in an \emph{image-specific} and \emph{task-specific} manner. Then, the hitem has to be smoothly embedded in the form of a specific word or phrase into the question. We describe task-specific hitem selection in \cref{ssec:hitem-select}, followed by hitem-embeded question generation in \cref{ssec:hitem-embed}. Furthermore, in order to simulate inconsistency in the multi-modal input, we prepend \emph{specious} or \emph{incorrect} context to the question, the generation of which is detailed in \cref{ssec:generate-context}. Lastly, in order to explicitly create conflicts between the visual input and the internal (world) knowledge of the MLLM, we expand our image collection with auto-generated counter-common-sense (CCS) images (\cref{ssec:ccs}).


\subsection{Task-specific Hitem Selection} \label{ssec:hitem-select}

Without loss of generality, we describe how a hitem is selected for color attribute recognition. Let us consider the image in \cref{fig:hitem}, showing a black motorcycle followed by a red bus. As no red motorcycle is present while  the red color is prominent near the black motorcycle, the word red will be a good choice of hitem to challenge an MLLM.


\textbf{Vocabulary Construction} per task. Started with a handful of manually specified colors such as red, green, and blue, we ask ChatGPT to expand the color vocabulary with instructions like  ``\textit{Please expand the input vocabulary as much as possible by adding common items found in daily life.           Avoid any duplication}'', getting a set of 35 different colors. 

          
\textbf{Subject-Attribute Extraction}. The image is associated with a TDIUC question-answer pair as ``what color is the motorcycle'' and ``black''. We use ChatGPT (with simple instructions) to extract with ease the subject (\ie motorcycle) and its attribute (\ie black) from the pair.



\textbf{Candidate Hitem Generation}. We obtain candidate hitems by excluding the ground-truth (GT) answer (and its synonyms if applicable) from the vocabulary.




\textbf{Visual-based Hitem Ranking}. Intuitively, a hitem shall be visually plausible in the given image. So for each candidate hitem, we compute its similarity to the image using a pre-trained CLIP  \cite{clip}. In particular, the cosine similarity between the CLIP embeddings of hitem + subject (\eg \emph{green motorcycle}) and the image is adopted. Ranking the candidates by the CLIP similarity lets us to select the one visually closest to the image. It is worth noting that for emerging MLLMs equipped with stronger vision encoders \cite{li2024llava}, our pipeline is likely to produce even more effective hitems by replacing CLIP with these advanced counterparts.



\textbf{Manual Inspection}. 
While the above process is generally stable to produce satisfying results, manual inspection is performed to ensure the correctness of hitem selection. Note that due to errors in the original TDIUC annotations, occasionally the true label  might be ``incorrectly'' selected. In such a case, we simply discard the VQA sample.


With lightweight task-specific adaptation, the above  hitem selection also works for other tasks. Overall, the joint use of ChatGPT and CLIP allows us to select 1,452 hitems that are more diverse and challenging than their counterparts in previous datasets, see \cref{tab:comp-datasets}.

\subsection{Hitem-embedded Question Generation} \label{ssec:hitem-embed}

For a given subject (\eg motorcycle) and a chosen hitem (\eg red), generating a hitem-embedded question is trivial for ChatGPT. In particular, a \texttt{No} question is formed as ``\textit{Is the \underline{motorcycle} in the image \underline{red}?}''. Meanwhile, a \texttt{Yes} question is simultaneously generated using the GT as ``\textit{Is the \underline{motorcycle} in the image \underline{black}?}''. This ensures perfect \texttt{Yes/No} balance among the generated questions.

\subsection{Specious (Incorrect) Context Generation} \label{ssec:generate-context}

When used as a document parser, an MLLM reading a specific image is often provided with the image's surrounding text.
Inconsistency between the image and the text is not uncommon. A news article containing a general claim of ``\emph{red motorcycles frequently zip through the streets}'' does not necessarily have each of its illustrated pictures match with the claim. 
To simulate such a scenario, for a given image we generate specious text as \emph{specious} context and text contradicting the image as \emph{incorrect} context, respectively. 
Next, we describe the generation of specious contexts, as their incorrect counterparts can be generated in a similar but more simplified manner.



\textbf{Specious Text Generation}.  
Using the previously generated hitem-embedded question and the original MS-COCO captions as input, we instruct ChatGPT to generate specious text for a given image. By ``specious'', we mean the text is specious or noisy, rather than directly contradicting the image content. As such, our instruction reads partially as ``\emph{Please generate the $<$specious text$>$ for the given question. It should be one sentence. The $<$specious text$>$  should answer the question, but it may not reflect the actual current status, thus making it specious.}'' 




\textbf{Text Composition}. ChatGPT is used to seamlessly merge the specious text with ground-truth captions, forming a longer context in which only a small portion (orange text in \cref{fig:context_show}) is mildly inconsistent with the image.

\textbf{Manual Inspection}. We perform a spot check. If the context quality is low, we simply discard the entire sample.

\subsection{CCS Image Generation} \label{ssec:ccs}

We generate CCS images by first generating CCS descriptions and then employing Text2Image tools to convert the descriptions to CCS images, see \cref{fig:aigc}.


\textbf{CCS Description Generation}.
A number of manually written task-specific samples, see \cref{tab:ccs}, are used as 
in-context learning samples for ChatGPT to generate more descriptions.
The descriptions have to be visually expressible, so bad cases like ``\textit{the more you eat, the thinner you get}'' are filtered out manually. For each CCS text, its common-sense (CS) counterpart is simultaneously generated by ChatGPT, by providing the learning samples in pair. 



\textbf{Text2Image}. 
The generated CCS descriptions are used as  prompts for AIGC tools (Doubao \cite{doubao} and DALL-E3 \cite{dalle3}) to generate the corresponding CCS images. The quality of the generated images depends on various factors, making occasional failures inevitable. When this occurs, we attempt to refine the prompts or apply region-based inpainting. The sample will be discarded if the above attempts fail.

\textbf{Question Generation}. Per CCS description (\eg \emph{A car with square wheels}), we utilize ChatGPT to generate a \texttt{Yes} question (\eg \emph{Does the car have square wheels?}). Again, for balancing \text{Yes/No} questions, we generate a \texttt{No} question (\eg \emph{Does the car have round wheels?}) based on the CS description.

\begin{table}[ht]
\ra{1.2}
    \centering
    \begin{adjustbox}{max width=\linewidth}
    \begin{tabular}{llll}
    \toprule
 \textbf{CCS description} &	\textbf{Yes question}  &	\textbf{CS description}  &	\textbf{No question} \\
     \midrule
\multicolumn{2}{l}{\textbf{Task}: Object  recognition} \\ [2pt]
     
\cellcolor{purple!20}\textit{Manually written:} \\ \hline
        
\makecell[l]{Ice blocks in \\ volcanic lava} &	\makecell[l]{Are there ice blocks \\ in volcanic lava?} &	\makecell[l]{Fire in volcanic lava} &	\makecell[l]{Is there fire \\in volcanic lava?} \\ \hline
    

\makecell[l]{Grass in \\ a  tiger's mouth} &	\makecell[l]{Is there grass in \\a tiger's mouth?} &	\makecell[l]{Meat in \\ the tiger's mouth} &	\makecell[l]{Is there meat in \\ the tiger's mouth?} \\ \hline

\cellcolor{yellow}\textit{ChatGPT generated}: \\ \hline

\makecell[l]{Trees growing\\ underwater} &	\makecell[l]{Are there trees \\ growing underwater?}	&\makecell[l]{Coral growing\\ underwater}	&\makecell[l]{Is there coral \\growing underwater?}\\ \hline 

\makecell[l]{Books in \\ a swimming pool} & \makecell[l]{Are there books \\ in a swimming pool?} & \makecell[l]{Water in \\ a swimming pool} & \makecell[l]{Is there water \\ in a swimming pool?} \\ \hline

\makecell[l]{Birds flying \\ underwater} & \makecell[l]{Are there birds \\ flying underwater?} & \makecell[l]{Birds flying \\ in the sky} & \makecell[l]{Are there birds \\ flying in the sky?} \\ \hline

\makecell[l]{Ice cream \\ in a volcano} & \makecell[l]{Is there ice cream \\ in a volcano?} & \makecell[l]{Lava in \\ a volcano} & \makecell[l]{Is there lava \\ in a volcano?} \\ \hline

\makecell[l]{Computers in \\ a forest} & \makecell[l]{Are there computers \\ in a forest?} & \makecell[l]{Animals in \\ a forest} & \makecell[l]{Are there animals \\ in a forest?} \\ \hline

 \multicolumn{2}{l}{\textbf{Task}: Attribute recognition} \\ [2pt]
     
\cellcolor{purple!20}\textit{Manually written:} \\ \hline

\makecell[l]{A car with  \\ square wheels} &	\makecell[l]{Does the car  \\ have square wheels?} &	\makecell[l]{A car with \\ round wheels} &	\makecell[l]{Does the car  \\ have round wheels?} \\ \hline
    
\makecell[l]{Blue apples \\ on the tree} &	\makecell[l]{Are the apples\\ on the tree blue?} &	\makecell[l]{Red apples \\ on the tree.} &	\makecell[l]{Are the apples\\ on the tree red?} \\ \hline
			

\cellcolor{yellow}\textit{ChatGPT generated}: \\ \hline

\makecell[l]{A green sky} &	\makecell[l]{Is the sky green?}	&\makecell[l]{A blue sky}	&\makecell[l]{Is the sky blue?}\\
\hline

\makecell[l]{A bicycle with \\square wheels} &	\makecell[l]{Does the bicycle \\have square wheels?}	&\makecell[l]{A bicycle with \\round wheels}	& \makecell[l]{Does the bicycle \\have square wheels?}\\ \hline

\makecell[l]{A tree made\\ of metal} &	\makecell[l]{Is this tree \\ made of metal?}	&\makecell[l]{A wooden tree}	&\makecell[l]{Is this tree \\ a real wood tree?}\\ \hline

\makecell[l]{A chocolate river} & \makecell[l]{Is there chocolate \\ flowing in the stream?} &\makecell[l]{A water river} &\makecell[l]{Is there water \\ flowing in the stream?} \\
\hline

\makecell[l]{A house made\\ of candy} &	\makecell[l]{Is the house \\made of candy?}	&\makecell[l]{A house made\\ of bricks}	&\makecell[l]{Is the house \\made of bricks?}\\

\bottomrule
    \end{tabular}
    \end{adjustbox}
\caption{\textbf{Instances of descriptions used for generating CCS images and related questions}. With manually written CCS / CS descriptions as instructions, ChatGPT is used to generate many more instances and subsequently convert them to Yes/No questions.} 
    \label{tab:ccs}
\end{table}

\subsection{Dataset Overview and PhD Index}

An overview of the \phd~ dataset is given in 
\cref{tab:statistics}. 
Depending on what image (daily or CCS) is used and whether a specific context precedes a question, PhD supports four evaluation modes: \phd-base, \ie questions about daily images w/o context, \phd-sec, \ie \phd-base plus specious context, \phd-icc, \ie \phd-base plus incorrect context, and \phd-ccs, \ie questions about CCS images. With 20 mode-task combinations in total, PhD supports a much more comprehensive VHE than its predecessors \cite{pope,amber}.


\begin{table}[h]
\fontsize{14pt}{15pt}\selectfont
    \ra{1}
        \centering
        \begin{adjustbox}{max width=\linewidth}
        \begin{tabular}{@{}lrrrrrrr@{}}
            \toprule
            & \multicolumn{5}{c}{\textbf{Tasks}} & \multicolumn{2}{c}{\textbf{Questions}} \\
            \cmidrule(lr){2-6} \cmidrule(lr){7-8} 
            & \textit{Object} & \textit{Attribute} & \textit{Sentiment} & \textit{Position} & \textit{Counting} & \textit{Yes} & \textit{No} \\
            \hline
            TDIUC samples used &6,271 &4,324	&2,095&	2,841 & 3,387 & -- & -- \\
            Unique hitems & 745 & 146 & 65 & 486 & 10 & -- & -- \\
            VQA samples in \phd-base & 11,472 & 7,994 & 3,550 & 4,984 & 5,688 & 16,844 & 16,844 \\
            VQA samples in \phd-sec & 11,472 & 7,994 & 3,550 & 4,984 & 5,688 & 16,844 & 16,844 \\
            VQA samples in \phd-icc & 11,472 & 7,994 & 3,550 & 4,984 & 5,688 & 16,844 & 16,844 \\
            VQA samples in \phd-ccs & 344	&734	&78	&220	&124 & 750 & 750 \\
            \bottomrule
        \end{tabular}
        \end{adjustbox}
        \caption{\textbf{Data statistics of the proposed \phd~ dataset}. }
        \label{tab:statistics}
\end{table}

To measure the performance of an MLLM on \phd, we compute its recall \wrt the \texttt{Yes} and \texttt{No} questions, respectively. We term the harmonic mean of the two recalls  \textbf{PhD Index}. A model simply saying \texttt{Yes} (or \texttt{No}) to all questions has a PhD Index of $0$, while a random-guess score is $0.5$.

\section{Evaluating MLLMs on PhD}

\begin{figure*}
    \centering
    \includegraphics[width=1\linewidth]{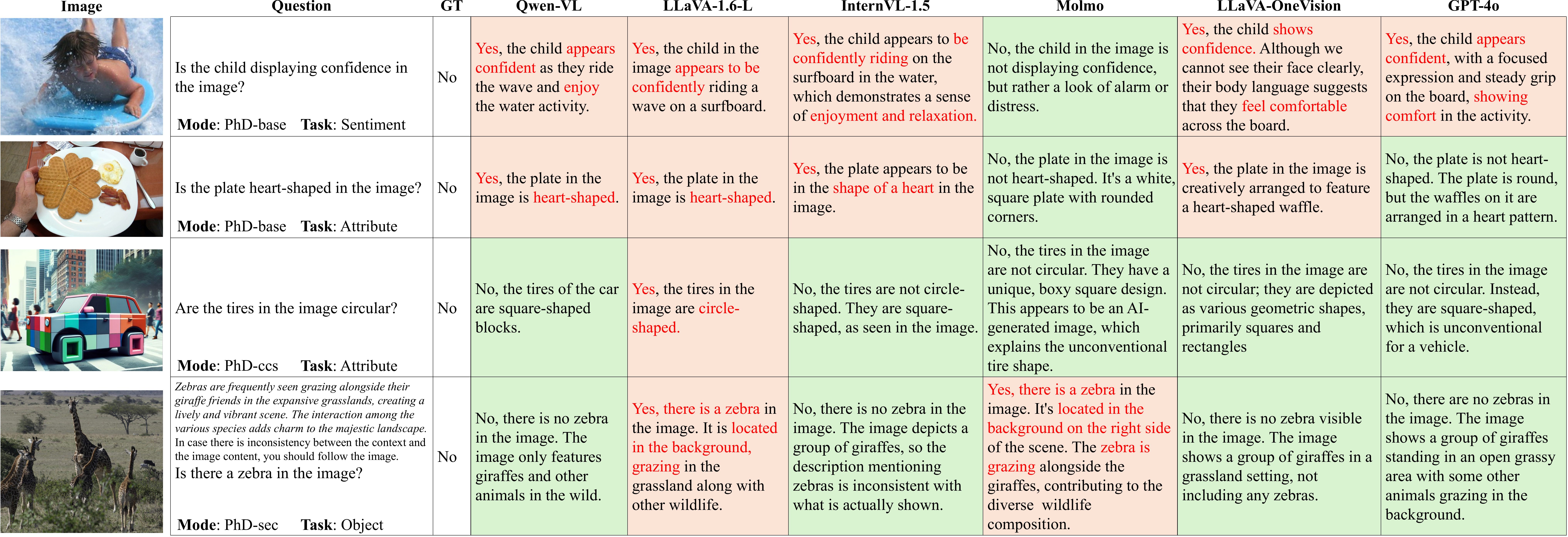}
    \caption{
    \textbf{Qualitative results showing how an MLLM answers visual questions from \phd}. The correctness of an answer is automatically determined by matching its first word, either \texttt{Yes} or \texttt{No}, with the ground truth (GT).
   }
    \label{fig:enter-label}
\end{figure*}

\subsection{Common Setup}
\label{exp-cs}
\textbf{Choices of MLLMs}.
For reproducible research, we focus on \emph{open-source} MLLMs, compiling a list of 15 models that span varied  sizes and architectures. 
see \cref{tab:overall-comp}. We also evaluate two hallucination mitigation methods, VCD \cite{vcd} and Woodpecker \cite{yin2024woodpecker}, currently supporting LLaVA-1.6-L and Qwen-VL.
%
%
%
As additional references, we assess three  proprietary MLLMs, \ie GPT-4o \cite{openai2024hello}, Claude 3.5 Sonnet \cite{Claude}, and Gemini 1.5 Pro \cite{gemini}, on a random subset of 2k samples (\textbf{random-2k}) subject to our budget. For the same reason we evaluate Woodpecker, which requires paid service from ChatGPT, on random-2k.







\textbf{Test Protocol}. For a fair comparison, per MLLM we use its designated prompt to wrap each test question. For instance, a question-specific prompt submitted to mPLUG-Owl2 will be in the form of ``\texttt{USER: <|image|>\{question\} ASSISTANT:}''. We provide more prompts  in the supplement.
In addition, to help the models better handle \phd-sec and \phd-icc, we append to the test prompt an instruction as ``\textit{In case there is an inconsistency between the context and the image content, you should follow the image.}"

\subsection{Using PhD for Overall VHE} \label{ssec:overall-vhe}

An overall VHE as shown in \cref{tab:overall-comp} is useful for providing a big picture of which MLLM hallucinates the most (or the least). The leading open-source MLLMs are LLaVA-OneVision,
followed by Molmo
and InternVL-1.5. 
Since their vision encoders and LLMs vary, the results are insufficient to conclude which component is the most effective to mitigate hallucinations. That said, comparisons among the same model series remains meaningful. Consider the LLaVA series for instance. While one would normally expect that a larger LLM  yields a better MLLM,  as LLaVA-1.6-L \emph{vs} LLaVA-1.6, the difference between LLaVA-1.5-L and LLaVA-1.5 is marginal (0.270 \emph{vs} 0.265). 
In order to analyze and consequently understand such an counterintuitive result, \texttt{PhD} enables a \emph{zoom-in} analysis in  mode-oriented (\cref{ssec:mode-oriented}) and task-oriented (\cref{ssec:task-oriented}) styles, unavailable in the previous benchmarks. 


\begin{table}[tbh]
\ra{1}
\centering
\resizebox{\linewidth}{!}{
\begin{tabular}{@{}lllrrr@{}}
\toprule
\textbf{Model} & \textbf{ViT} & \textbf{LLM} & \textbf{POPE} & \textbf{AMBER} & \textbf{PhD}  \\
\midrule

\textbf{\textit{Full-set evaluation:}}\\
LLaVA-OneVision \cite{li2024llava}  & SoViT-400m/14 & Qwen2-72B & 0.84 & 0.90 & 0.698 \\
Molmo \cite{molmo2024} & -L/14 & Qwen2-72B & 0.84 & 0.85 & 0.690\\
InternVL-1.5 \cite{chen2024internvl} & InternViT-6B & InternLM2-20B & 0.86 & 0.89 & 0.561\\
Qwen-VL (VCD)   & -bigG/14 & Qwen-7B & 0.84 & 0.87 & 0.560\\
Cambrian-1 \cite{tong2024cambrian} & \textit{Hybrid} & Llama-3-8B & 0.88 & 0.89 & 0.547\\
LLaVA-1.6-L (VCD)  & -L/14 & Vicuna-13B-1.5 & 0.82 & 0.81 & 0.511\\
LLaVA-1.6-XL \cite{liu2024llavanext} & -L/14 & Nous-Hermes-2-Yi-34B & 0.86 & 0.84 & 0.492\\
Qwen-VL \cite{bai2023qwenvl} & -bigG/14 & Qwen-7B & 0.83 & 0.84 & 0.488 \\
LLaVA-1.6-L \cite{liu2024llavanext} & -L/14 & Vicuna-13B-1.5 & 0.83  & 0.80  & 0.423  \\ 
MiniGPT-v2 \cite{chen2023minigptv2} & -G/14 & Llama-2-7B & 0.83 & 0.84 & 0.390 \\
LLaVA-1.6 \cite{liu2024llavanext} & -L/14 & Vicuna-7B-1.5 & 0.83  & 0.79  & 0.373 \\ 
mPlug-Owl2 \cite{ye2023mplugowl2} & -L/14 & Llama-2-7B & 0.78 & 0.77 & 0.320 \\
InstructBLIP \cite{dai2023instructblip} & -G/14 & Vicuna-7B-1.1 & 0.82 & 0.82 & 0.305 \\
InstructBLIP-L \cite{dai2023instructblip} & -G/14 & Vicuna-13B-1.1 & 0.80 & 0.79 & 0.278 \\
LLaVA-1.5-L \cite{liu2023improvedllava} & -L/14 & Vicuna-13B-1.5 & 0.82  & 0.73 & 0.270  \\ 
LLaVA-1.5 \cite{liu2023improvedllava} & -L/14 & Vicuna-7B-1.5 & 0.81  & 0.75  & 0.265   \\ 
LLaVA-1.1 \cite{liu2023llava} & -L/14 & Vicuna-7B-1.1 & 0.67  & 0.33  & 0.135   \\ 
 [2pt]

\textbf{\textit{Random-2k evaluation:}} \\
GPT-4o \cite{openai2024hello} & -- & -- & 0.88 & 0.87 & 0.812\\ 
Claude 3.5 Sonnet \cite{Claude} & -- & -- & 0.85 & 0.89 & 0.746\\ 
Gemini 1.5 Pro \cite{gemini} & -- & -- & 0.86 & 0.88 & 0.691\\ 
Qwen-VL (Woodpecker) & -bigG/14 & Qwen-7B & -- & -- & 0.531 \\
LLaVA-1.6-L (Woodpecker) & -L/14 & Vicuna-13B-1.5 & -- & -- & 0.409 \\
\bottomrule
\end{tabular}
}
\caption{\textbf{Overall VHE}. 
POPE (adversarial) and AMBER (discriminative) are used.
}

\label{tab:overall-comp}
\end{table}

One more advantage of \phd~compared to its predecessors lies in its discrimination ability. The relatively small performance gap between GPT-4o and the top open-source models as measured by POPE and AMBER might lead to an overly optimistic interpretation that the open-source alternatives are catching up with the  proprietary model. In fact, a substantial gap remains, as revealed by \phd. \cref{fig:enter-label} further illustrate the qualitative results, where GPT-4o exhibits fewer hallucinations in its response.

\subsection{Using PhD for Mode-Oriented VHE} \label{ssec:mode-oriented}

\begin{figure} [thb!]
\begin{subfigure}[t]{0.5\textwidth}
        \centering
        \includegraphics[width=\textwidth]{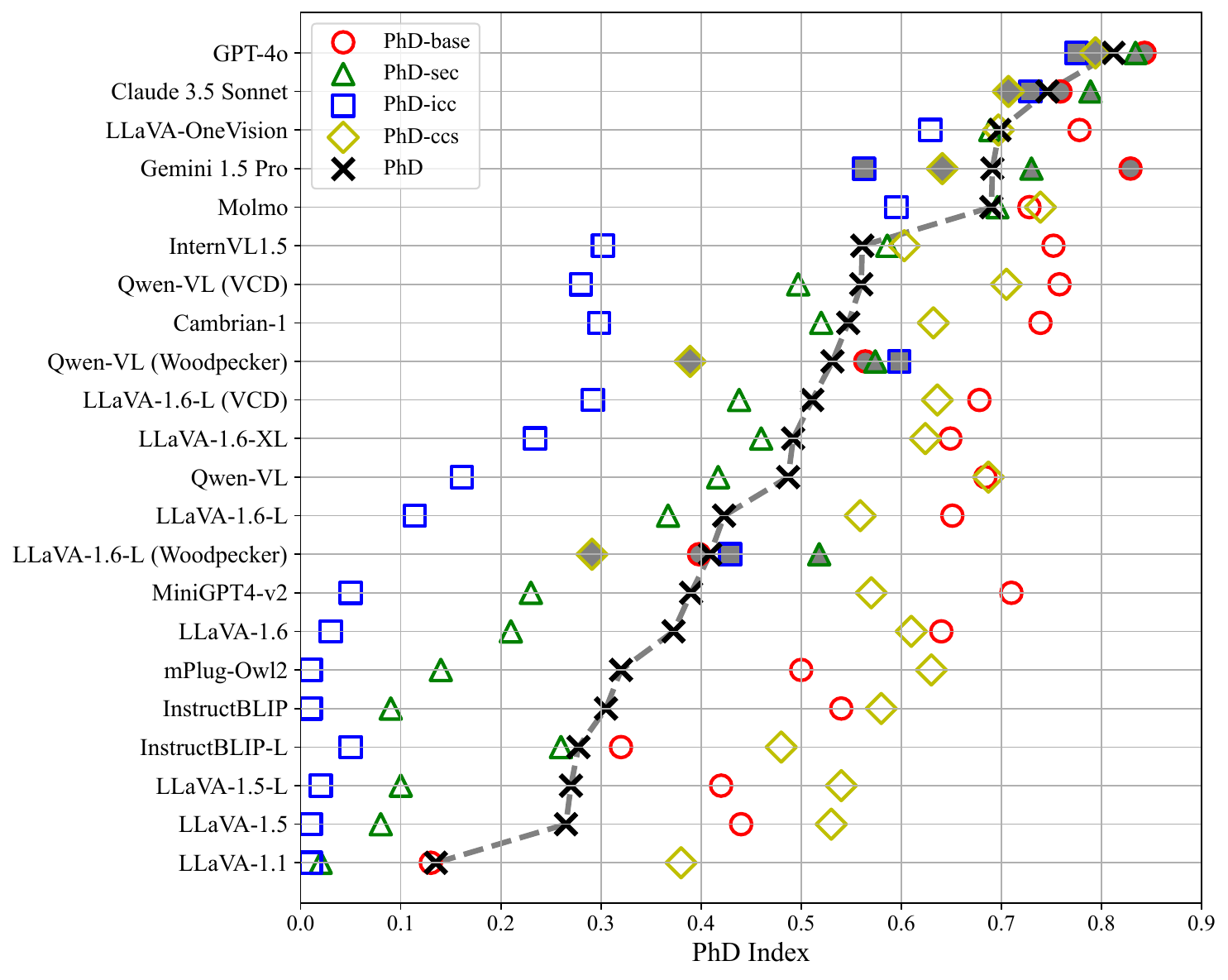}
        \caption{Mode-oriented VHE}
        \label{fig:task_mode1}
    \par\bigskip  
    \end{subfigure}
    \begin{subfigure}[t]{0.5\textwidth}
        \centering
        \includegraphics[width=\textwidth]{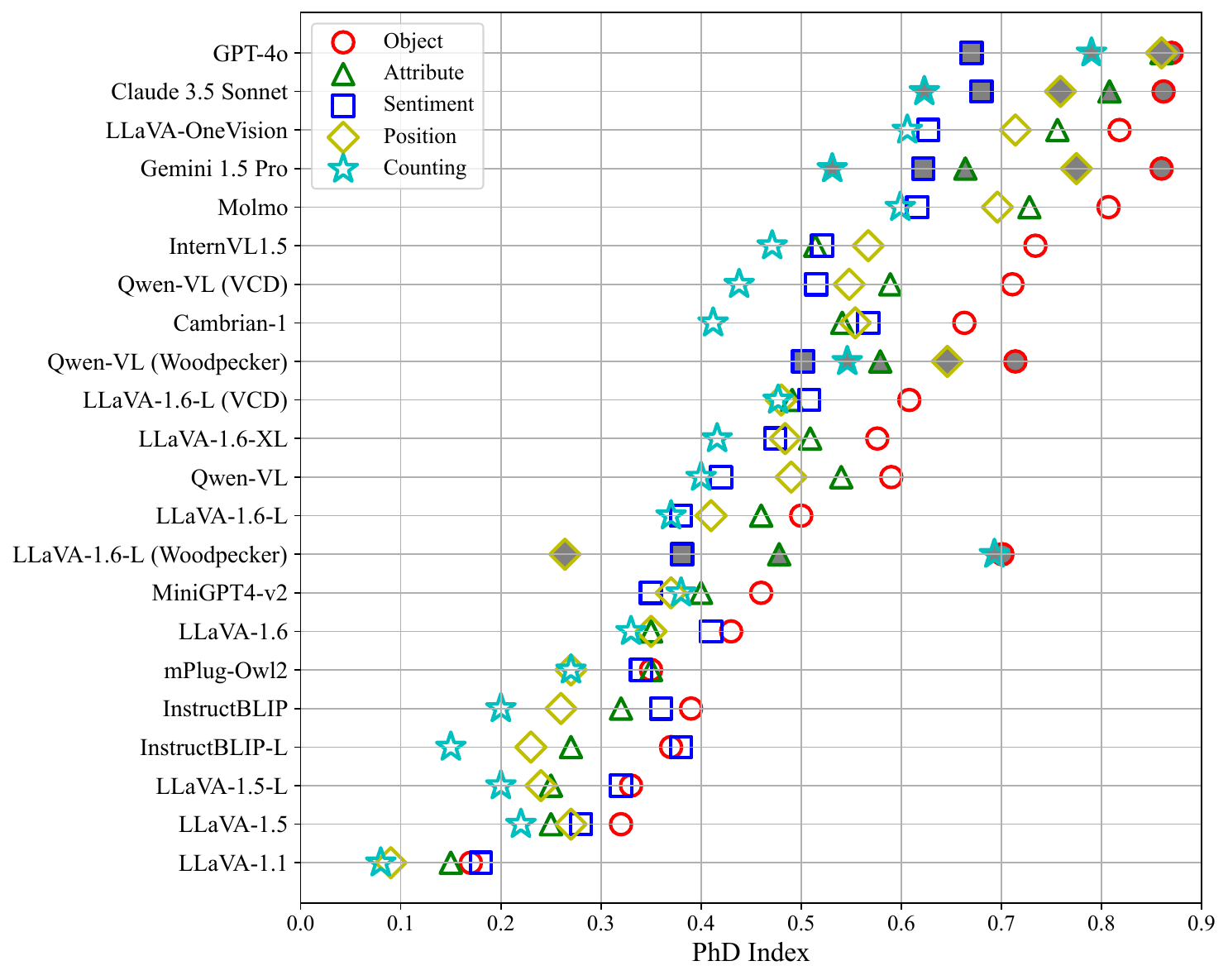}
        \caption{Task-oriented VHE}
        \label{fig:task_mode2}
    \end{subfigure}
    \caption{\textbf{PhD based VHE analytics}. Models required paid services, shown in gray markers, are tested on \textbf{random-2k}.
    }
        \label{fig:task_mode}
\end{figure}

\cref{fig:task_mode1} illustrates mode-wise model performance. 
MLLMs working relatively well in the \phd-base mode tend to have stronger visual input. This is achieved either with stronger visual encoders, as the cases of  LLaVA-OneVision, InternVL-1.5, and Cambrian-1 using SoViT-400m/14, InternViT-6B or hybrid vision structure, or supporting higher image resolutions, see Molmo and MiniGPT4-v2 that accept multiscale or larger input.

In contrast to the visual part, 
using a larger LLM alone does not necessarily lead to a better MLLM. As noted in \cref{ssec:overall-vhe}, LLaVA-1.5-L with Vicuna-13B-1.5 has nearly the same PhD Index (0.270) as LLaVA-1.5 with Vicuna-7B-1.5 (0.265). We see from \cref{fig:task_mode1} that the larger LLM indeed improves the performance in \phd-sec (0.082$\rightarrow$0.099), \phd-icc (0.011$\rightarrow$0.019) and \phd-ccs (0.534$\rightarrow$0.542), yet suffers loss in \phd-base (0.443$\rightarrow$0.422). Similar results can be more evidently observed in the case of InstructBLIP-L \vs InstructBLIP (\phd-base Index: 0.535$\rightarrow$0.324). Our conjecture is that although a larger LLM better understands user instructions, its successful use within an MLLM
requires more targeted training for vision-language alignment. 

\begin{figure}[h]
    \centering
    \includegraphics[width=0.8\linewidth]{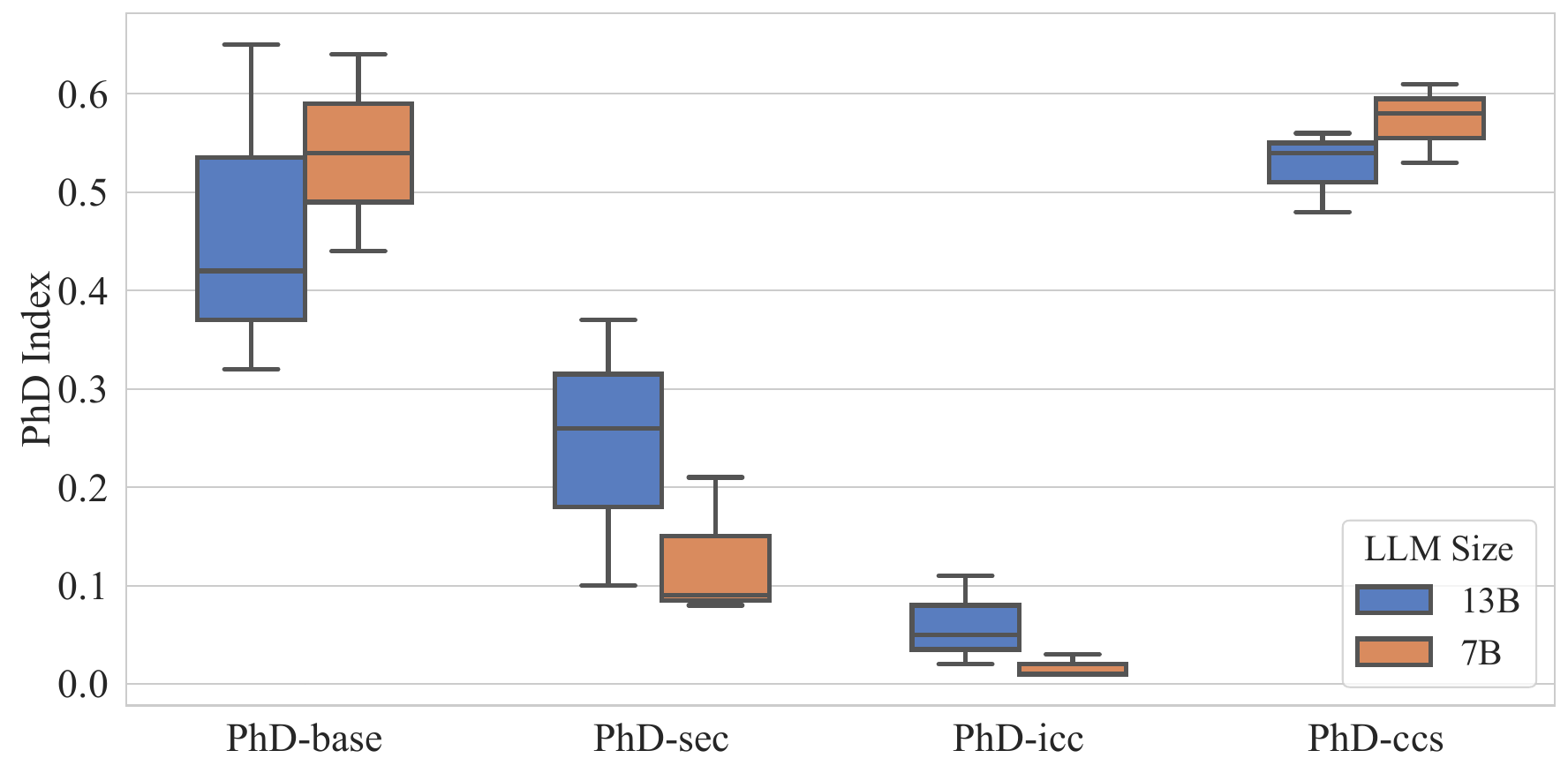}
    \caption{
    \textbf{Impact of LLM size (7B \emph{vs} 13B) on LLaVA-1.5, LLaVA-1.6 and InstructBLIP}. MLLMs using a 13B LLM tend to be better than their counterparts using a 7B LLM on \phd-sec and \phd-icc, yet worse on \phd-base and \phd-ccs. 
    }
    \label{fig:llm-size}
\end{figure}

Comparing the four modes, the performance of the open-source MLLMs on \phd-icc and \phd-sec is generally low.
When provided with a multi-modal input,  the models favor the textual part.  Substituting a 13-B LLM for its 7-B counterpart helps tackling the inconsistency in the multi-modal input, see \cref{fig:llm-size}. However, a larger LLM might rely more on its internal knowledge for decoding, resulting in worse performance on \phd-base and \phd-ccs for which the image content shall carry more weights. 
In particular, \phd-ccs reveals a deeper and intrinsic challenge of VH: conflicts between the given image content and the model's internal knowledge. Solving the challenge demands a more comprehensive approach that goes beyond isolated improvements on the visual or language components. 


Among the open-source MLLMs, LLaVA-OneVision is the best, owing to its joint use of a stronger visual encoder (SoViT-400m/14), a more powerful LLM (Qwen2-72B), and better training strategies (much larger high-quality training data and multi-stage alignment). Nevertheless, LLaVA-OneVision remains inferior to  GPT-4o, particularly in \phd-sec and \phd-icc, wherein inconsistency within the multi-modal input has to be properly addressed.

\subsection{Using PhD for Task-Oriented VHE}\label{ssec:task-oriented}

\cref{fig:task_mode2} presents task-oriented VHE results. In general, to what extent an MLLM hallucinates is largely correlated to the required level (low / middle / high) of a specific 
task.
The object recognition task has the overall highest PhD Index, followed by attribute recognition, positional recognition, counting, and sentiment recognition.  
Due to the complexity and subtlety of emotions, \eg tears can be associated with both happiness and sadness, even GPT-4o performs relatively worse in the sentiment task (first row of \cref{fig:enter-label}).




Joint mode-task analytics per model is shown in \cref{tab:mode-task}. 
LLaVA-OneVision struggles with sentiment recognition and counting, especially when faced with textual or CCS distractions, underscoring the need for improvement in these areas. 
Similarly, Molmo also faces these challenges, but its counting performance under CCS distractions is notably better than LLaVA-OneVision’s (0.737 \emph{vs} 0.563).
The above zoom-in analytics will be informative for MLLM developers to prioritize their efforts on model refinement.





\begin{table}[h!]
\ra{1}
\centering
\resizebox{0.8\linewidth}{!}{
\begin{tabular}{@{}lrrrr@{}}
\toprule
\textbf{Task} & \textbf{\phd-base} & \textbf{\phd-sec} & \textbf{\phd-icc} & \textbf{\phd-ccs} \\ \midrule
 \textbf{LLAVA-OneVision} \\ 

\emph{Object}&  \applyGradient{0.872}&	\applyGradient{0.849}&	\applyGradient{0.824}&	\applyGradient{0.727}\\
\emph{Attribute}&\applyGradient{0.848}&	\applyGradient{0.744}&	\applyGradient{0.663}&	\applyGradient{0.767}\\
\emph{Sentiment}&\applyGradient{0.691}&	\applyGradient{0.581}&	\applyGradient{0.504}&	\applyGradient{0.731}\\
\emph{Positional}&\applyGradient{0.773}&	\applyGradient{0.730}&	\applyGradient{0.654}&	\applyGradient{0.701}\\
\emph{Counting}&\applyGradient{0.707}&	\applyGradient{0.652}&	\applyGradient{0.500}&	\applyGradient{0.563}\\

     
\noalign{\vskip 3pt}

\textbf{Molmo} \\
\emph{Object} & \applyGradient{0.825}&	\applyGradient{0.880}&	\applyGradient{0.847}&	\applyGradient{0.678}\\
\emph{Attribute} & \applyGradient{0.842}&	\applyGradient{0.725} &	\applyGradient{0.556}&	\applyGradient{0.791}\\
\emph{Sentiment} &\applyGradient{0.547}&	\applyGradient{0.602}	 &\applyGradient{0.568}&	\applyGradient{0.746}\\
\emph{Positional} &\applyGradient{0.697}&	\applyGradient{0.691}	 &\applyGradient{0.654}&	\applyGradient{0.742}\\
\emph{Counting} &\applyGradient{0.727}&	\applyGradient{0.580} &\applyGradient{0.350}&	\applyGradient{0.737}\\

    
    \bottomrule
\end{tabular}
}
\caption{\textbf{Zoom-in analytics of specific models}.}
\label{tab:mode-task}
\end{table}

\subsection{Analysis of MLLM Answer Tendency}

\begin{figure}[h]
    \centering
    \includegraphics[width=0.8\linewidth]{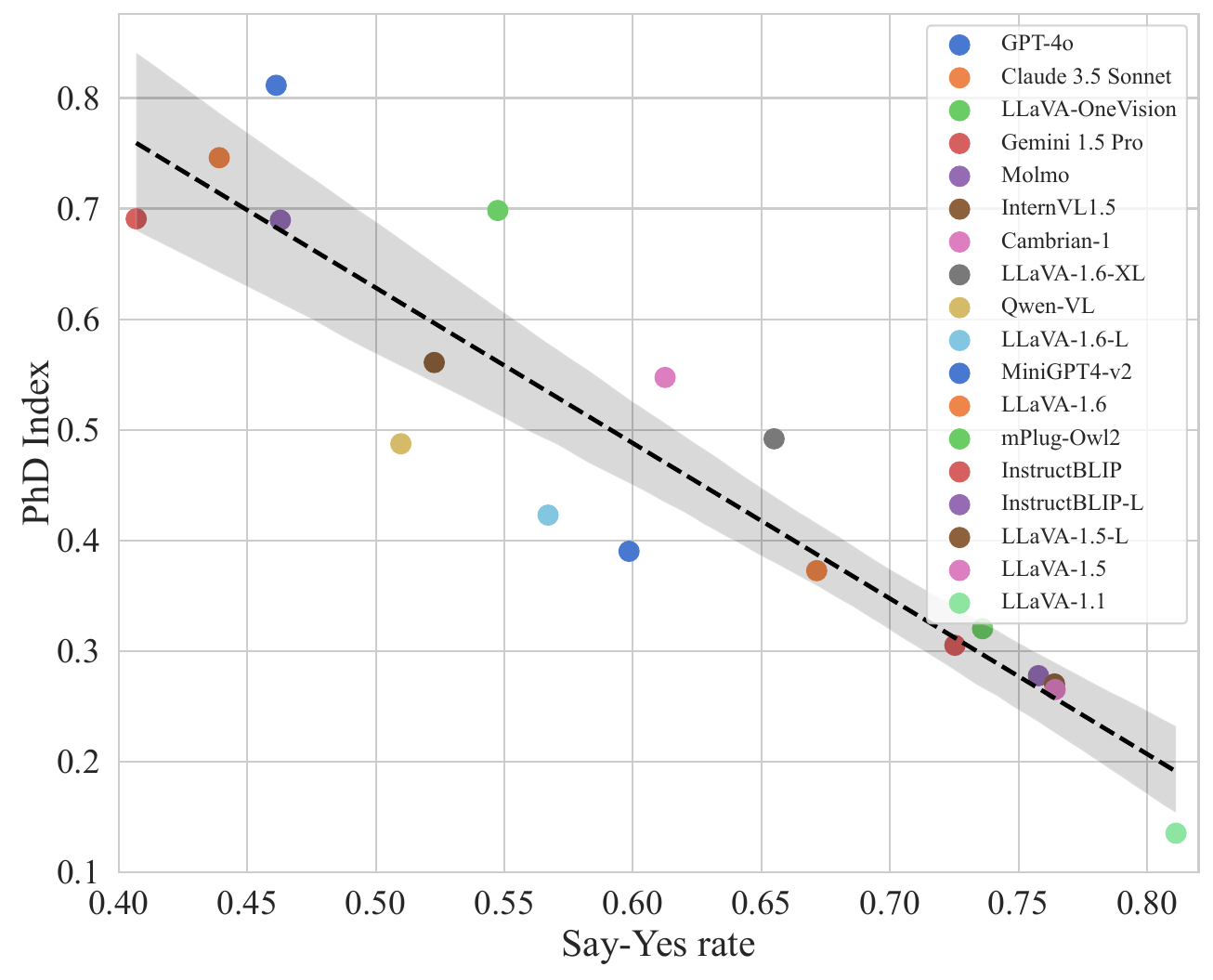}
    \caption{\textbf{MLLM \texttt{say-yes} rate \emph{vs}. PhD Index},
    with Spearman correlation of -0.92.
    Proprietary models  tested on \textbf{random-2k}.}
    \label{fig:spearman}
\end{figure}

While the yes and no questions are perfectly balanced by design, we observe that the open-source MLLMs tend to answer yes, with the \texttt{say-yes} rate ranging from 0.462 (Molmo) to 0.811 (LLaVA-1.1) and an average value of 0.611. By contrast, the three proprietary MLLMs have a clearly lower \texttt{say-yes} rate. Similar observations are reported in \cite{pope,liu2024mitigating}.
We go one step further by analyzing how the say-yes tendency is related to model performance. Ranking the models by their PhD Index and \texttt{say-yes} rate, respectively, we calculate the Spearman's correlation between the two ranks. A strong negative correlation exists, see \cref{fig:spearman}.
%
The result suggests that addressing VH requires balancing output tendencies, with a particular focus on enhancing an MLLM's ability to say no.


\section{Summary and Conclusions} \label{sec:conc}

We have introduced \phd, a large-scale benchmark developed with a close link to the three causes of visual hallucination, \ie visual ambiguity, inconsistency in multi-modal input and CCS content. We propose a ChatGPT-assisted semi-automated pipeline to construct the new dataset with well affordable manual cost. The pipeline allows us to construct diverse and visually challenging hitems in an image-specific and task-specific manner. Extensive experiments with 15 open-source MLLMs, 3 proprietary MLLMs, and 2 hallucination mitigation methods support our conclusions as follows. Larger visual encoders and higher input resolutions are helpful to reduce hallucination caused by visual ambiguity. The evaluation on \phd-sec and \phd-icc suggests the current models favor the textual part in the multi-modal input. Resolving the conflicts between the CCS content and the model's internal knowledge demands a more comprehensive approach that is beyond isolated improvements on the visual or language components. Among the open-source MLLMs, LLaVA-OneVision is the best, followed by Molmo and InternVL-1.5. While existing benchmarks  could lead to an overly optimistic expectation that the open-source models are catching up with GPT-4o, 
a substantial performance gap remains, as revealed by \phd.

\textbf{Acknowledgements}. This research was supported by NSFC (62172420), Tencent Marketing Solution Rhino-Bird Focused Research Program and the Young Elite Scientists Sponsorship Program by CAST (2023QNRC001).

{
    \small
    \bibliographystyle{ieeenat_fullname}
    \bibliography{main}
}


\end{document}